\documentclass[10pt, a4paper]{article}

\usepackage{lrec-coling2024}  

\usepackage{tikz}
\usetikzlibrary{positioning}
\usepackage{svg}
\usepackage{csquotes}
\usepackage{pifont}

\usepackage{xstring}

\usepackage{color}

\title{CASIMIR: A Corpus of Scientific Articles enhanced with Multiple Author-Integrated Revisions}

\name{Léane Jourdan$^1$, Florian Boudin$^{1,2}$, Nicolas Hernandez$^1$, Richard Dufour$^1$} 

\address{$^1$Nantes Université, École Centrale Nantes, CNRS, LS2N, UMR 6004, F-44000 Nantes, France \\
$^2$Japanese French Laboratory for Informatics, CNRS, NII, Tokyo, Japan\\
\{Leane.Jourdan,Florian.Boudin, Nicolas.Hernandez,Richard.Dufour\}@univ-nantes.fr\\}

\abstract{
Writing a scientific article is a challenging task as it is a highly codified and specific genre, consequently proficiency in written communication is essential for effectively conveying research findings and ideas.
In this article, we propose an original textual resource on the revision step of the writing process of scientific articles.
This new dataset, called \texttt{CASIMIR}, contains the multiple revised versions of 15,646 scientific articles from OpenReview, along with their peer reviews. Pairs of consecutive versions of an article are aligned at sentence-level while keeping paragraph location information as metadata for supporting future revision studies at the discourse level.
Each pair of revised sentences is enriched with automatically extracted edits and associated revision intention.
To assess the initial quality on the dataset, we conducted a qualitative study of several state-of-the-art text revision approaches and compared various evaluation metrics.
Our experiments led us to question the relevance of the current evaluation methods for the
text revision task.
 \\ \newline  }

\begin{document}

\maketitleabstract

\section{Introduction}
Writing a scientific article is a complex and challenging task, especially for young researchers who need to learn the conventions of scientific writing or non-native English-speaking researchers who also have to overcome the language barrier. Whether junior or senior, all researchers must pay attention to the quality of their writing in order to effectively convey their ideas to the reader. 
The difficulties result from scientific writing being a genre with its own conventions and specificities, including the structure of the article (e.g., IMRaD format: \textit{Introduction, Methods, Results and Discussion}~\citep{swales_genre_analysis}), a concise and precise style, the use of tenses, pronouns, or terminology~\citep{kallestinova2011write,bourekkache2022english}.
Various aspects of scientific writing assistance have been explored, including text revision~\citep{du-etal-2022-read}, spell checking or predicting paper acceptance/rejection~\citep{kang-etal-2018-dataset}.

Corpora comprising multiple versions of revised scientific articles are essential as they enable in-depth analysis of the iterative revision process undertaken to achieve a satisfying research paper. Such datasets are invaluable for training automated systems designed to assist in scientific writing. However, the few existing corpora may have limitations such as insufficient size for comprehensive training, incomplete articles, limited context, only two versions of articles (i.e. intermediate versions are excluded), or absence of associated reviews.

In this article, we introduce a new dataset, CASIMIR\footnote{\href{https://huggingface.co/datasets/taln-ls2n/CASIMIR}{https://huggingface.co/datasets/taln-ls2n/CASIMIR}}, composed of multiple versions of 15,646 full-length scientific articles in English collected from OpenReview\footnote{\href{https://openreview.net}{https://openreview.net}}. This platform offers less-finalized initial versions, thus resulting in more substantial revisions. The dataset includes 3.7 million pairs of automatically aligned edited sentences, representing 5.2 million of individual edits, each annotated with an automatic revision intention labeling tool (e.g.,~adding content, fixing grammar). Each paper is supplemented with metadata, including associated peer reviews and venue information.

Our approach to constructing the dataset draws inspiration from the work of~\citet{du-etal-2022-understanding-iterative} and~\citet{jiang-etal-2022-arXivEdits}, which involves collecting and aligning multiple versions of a single paper.
Our dataset distinguishes itself from existing ones in two significant ways: firstly, its size is an order of magnitude larger; and secondly, it offers both sentence-level alignment and paragraph-level localization information, providing support for the development of future discourse-level revision tools.
To get a better understanding of the quality of our dataset, we conducted a qualitative analysis and evaluated the performance of several state-of-the-art text revision models.

Our contributions are as follows:
\begin{enumerate}
    \item We released a large and open corpus freely available to the research community for revision in scientific articles. 
    \item We conducted a qualitative analysis of the content of this corpus.
    \item We evaluated three models on the task of sentence text revision and compared various metrics to evaluate this task.
\end{enumerate}

\section{Related Work}
\label{sec:append-how-prod}

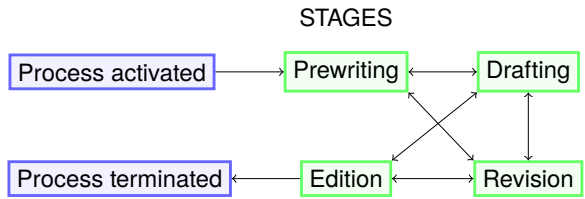
\begin{figure}
\centering
\resizebox{.49\textwidth}{!}{
\begin{tikzpicture}[
roundnode/.style={circle, draw=green!60, fill=green!5, very thick, minimum size=7mm},
titlenode/.style={rectangle, draw=white!60, fill=white!5, very thick, minimum size=5mm},
bluerectangle/.style={rectangle, draw=blue!60, fill=blue!5, very thick, minimum size=5mm},
greenrectangle/.style={rectangle, draw=green!60, fill=green!5, very thick, minimum size=5mm}
]
\node[greenrectangle]      (step1)                              {Prewriting};
\node[greenrectangle]        (step4)       [below=of step1] {Edition};
\node[greenrectangle]      (step2)       [right=of step1] {Drafting};
\node[greenrectangle]        (step3)       [below=of step2] {Revision};
\node[bluerectangle]        (start)       [left=of step1] {Process activated};
\node[bluerectangle]        (finish)       [left=of step4] {Process terminated};
\node[titlenode, yshift=-22pt]        (title)       [above=of step1] {STAGES};
\draw[->] (start.east) -- (step1.west);
\draw[<->] (step1.east) -- (step2.west);
\draw[<->] (step2.south) -- (step3.north);
\draw[<->] (step4.east) -- (step3.west);
\draw[<->] (step2.south west) -- (step4.north east);
\draw[<->] (step1.south east) -- (step3.north west);
\draw[<-] (finish.east) -- (step4.west);
\end{tikzpicture}
}
\caption{The writing process of a scientific article, inspired by~\citet{seow2002writing}} \label{fig:process_2}
\end{figure}

\paragraph{Scientific writing process} Previous works~\citep{silveira2022guide,laksmi2006scaffolding,bailey2014academic,seow2002writing,du-etal-2022-read,jourdan2023text} described the writing of a scientific paper as a four-step process, as illustrated in Figure~\ref{fig:process_2}. %
Those four steps are: 1: Prewriting (collecting and organizing ideas, writing the outline), 2: Drafting (writing full sentences from notes and focusing on content rather than form and structure), 3: Revision (changing the structure of paragraphs and content of sentences, focusing on conciseness, clarity, connecting elements, and simplifying the text) and 4: Edition (spelling error correction, minor changes, and editing figures and tables). %

Our corpus targets the \textit{Revision} step, which is characterized by substantial alterations to the text, including changes to content, sentence structure, and the logical flow of ideas. 
Text revision is an iterative task that often involves multiple iterations until the structure and phrasing is satisfying~\citep{du-etal-2022-read}. It is also 1-to-N, as one segment of text can have multiple correct revisions~\citep{ito-etal-2019-diamonds}. Providing automated assistance at the revision step of the writing process could enable authors to efficiently improve their writing.
To train and evaluate such scientific writing assistance tools, some corpora are needed.
While existing corpora for general domain text revision are typically gathered from Wikipedia by collecting the pages' history of revisions~\citep{yang-etal-2017-identifying-semantic,faruqui-etal-2018-wikiatomicedits,wu-etal-2021-automatic,dwivediyu2022editeval},
our research is dedicated to resources focusing on scientific writing.%

\paragraph{Revision datasets in the scientific domain} Datasets for text revision composed of scientific papers vary in their content and scale. Some datasets only encompass the title, abstract, and introduction of scientific papers ~\citep{du-etal-2022-understanding-iterative,mita2022automated} or isolated sentences~\citep{ito-etal-2019-diamonds}. 
Others are relatively small in size, making them unsuitable for proper training of tools based on Language Models for the text revision task~\citep{jiang-etal-2022-arXivEdits,darcy2023aries,10.1162/coli_a_00455}. However, these smaller datasets can still serve as valuable resources for model evaluation, and they can be combined with each other for training. 

In the studies conducted by~\citet{du-etal-2022-understanding-iterative,jiang-etal-2022-arXivEdits,ito-etal-2019-diamonds}, the focus was primarily centered on sentence-level alignments. Nonetheless, retaining information about the structural organization of paragraphs in scientific articles can enable the consideration of a coherent broader context in revision models.

Some of these datasets do not contain associated peer reviews~\citep{du-etal-2022-understanding-iterative,jiang-etal-2022-arXivEdits}. Furthermore, datasets designed for predicting paper acceptance or rejection, like the one presented in~\citet{kang-etal-2018-dataset}, typically offer only a single version of the paper, as they were not originally created for the specific task of text revision. A limitation regarding the number of revisions also exists with ARIES~\citep{darcy2023aries}. ARIES is the most closely related resource to our work. It is also collected from OpenReview and includes complete documents along with peer reviews. However, it was primarily constructed for the edit-review alignment task, providing only two versions (the initial submission and the final version) for each article, despite many papers submitted to OpenReview having multiple versions. The current version of the ARIES dataset contains a relatively modest collection of 1,720 research papers.

The CASIMIR corpus aims to offer a large resource for training models, with multiple versions of full-length scientific articles and associated reviews.

\section{Corpus Creation}
\begin{figure*}
    \centering
    \includegraphics[width=\textwidth]{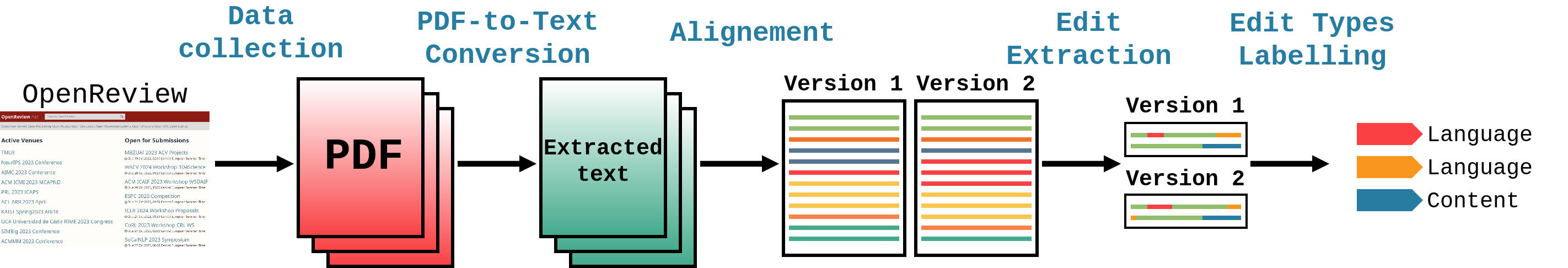 }
    \caption{Steps of the creation process of the CASIMIR corpus}
    \label{fig:creation_process}
\end{figure*}

This section outlines the creation process of the CASIMIR corpus summarized in Figure~\ref{fig:creation_process}. A manual qualitative evaluation of steps 3 and 4 was also conducted on a sample (369 sentences from 6 pairs of articles) in order to validate the quality of our corpus.

\subsection{Large Data Collection} 

OpenReview is an open platform for peer review that allows hosting different versions of the same article in PDF along with their reviews. It offers less-finalized initial versions, thus resulting in more substantial revisions. %
Furthermore, the peer reviews and authors' replies are directly posted on a dedicated forum space for each article. The content of the posts can serve as a guide to the quality of associated articles and the underlying intentions behind the revisions made.
However, OpenReview only offers PDF version of the papers with no associated LaTeX file, thus requiring textual content extraction.
 We collected all available documents on OpenReview as of March 10, 2023. 
\subsection{PDF-to-Text Conversion}
\label{sec:grobid}
Several tools exist to extract the textual content of PDF files while preserving their structure, such as GROBID~\citep{GROBID}, a well-known tool for its quality of text extraction. However, after conducting an initial assessment of the conversion quality on a subset of documents, it exhibited errors such as incorrect table and figure detection, partial sentence removal, improper identification of paragraphs as figures and their shift to the end of the document. These errors make alignment between different versions of articles too complex.\\%
Finally, we rather employed the VILA tool~\citep{shen-etal-2022-vila} that gives satisfactory results. Note that using this tool, some PDF conversion problems remain, such as inaccurate section detection, and transcription of formulas included within paragraphs.
However, all the content is kept, and the text order is maintained.

After conversion, the bibliography is removed, and equations, figures and tables replaced by tags (\texttt{[Equation]},\texttt{[Figure]} and \texttt{[Table]} respectively). 
We also split the text data in paragraphs and cleaned it from page numbers, line numbers, and line breaks with rule-based heuristics. PDF files that cannot be converted are excluded from the corpus.

\subsection{Alignment and Edit Extraction}
For each pair of two consecutive versions of a paper, we aligned the textual content at sentence-level and extracted the edits at word-level. First, we performed sentence-level alignment using Bertalign~\citep{bertalign} that was found to perform very well on the sentence alignment task.~\citep{bertalign} report a performance of 99\% on the alignment task, after manual evaluation we obtained a micro-accuracy of 89,70\%.%

Then, from each pair of aligned sentences, we extracted the edits between the two versions at word-level using \href{https://git-scm.com/docs/git-diff}{\texttt{git-diff}}\footnote{\href{https://git-scm.com/docs/git-diff}{https://git-scm.com/docs/git-diff}}. 

\subsection{Edit Types Labelling}
We automatically annotated the extracted edits with a revision intention.
For this step of the creation process, we used the \href{https://huggingface.co/chaojiang06/arXivEdits-intention-classifier-T5-large-coarse}{arXivEdits intention classifier}\footnote{\href{https://huggingface.co/chaojiang06/arXivEdits-intention-classifier-T5-large-coarse}{https://huggingface.co/chaojiang06/arXivEdits-intention-classifier-T5-large-coarse}}. \citealp{jiang-etal-2022-arXivEdits} report an accuracy of 84.4\% for their coarse version leading us to use this classifier instead of their fine-grained version or the most frequently used classifier from~\citealp{du-etal-2022-understanding-iterative}. After manual evaluation we obtained a micro-accuracy of 80,63\%.

As defined by~\citealp{jiang-etal-2022-arXivEdits}, the generated labels are as follows :%
\begin{itemize}
    \item \textbf{Content}: \textquote{Update large amount of scientific content, add or delete major fact.}
        
    \item \textbf{Improve-grammar-Typo}: \textquote{Fix grammatical errors, correct typos, or smooth out grammar needed by other changes.}
    \item \textbf{Format}: \textquote{Adjust table, figure, equation, reference, citation, and punctuation etc.}
    \item \textbf{Language}: Adjust language to make the text more accurate, coherent, professionally sounding and improve its readability.
\end{itemize}

\subsection{Corpus split}
Finally, we divided our dataset into three parts: 80\% for training, 10\% for validation, and another 10\% for testing. Additionally, we offer a smaller test dataset as a subset of the larger one. This smaller test dataset accounts for 30\% of the test split (i.e. 3\% of the corpus), mainly because running inference on large models using the large test set could be time-consuming and resource-intensive.

\section{Corpus Analysis}
In this section, we conducted a qualitative analysis on the content of our dataset. 
We began by studying the distribution of the number of versions and reviews by article. Then, we investigated the distribution of edits, examining both their quantity and types within the versions. Lastly, we examined how these edits change over time and where they are found within the articles. This analysis provides insights into the dataset's content, offering an understanding of the revision process.

We studied the distributions of the number of versions and reviews by article, the number of edits by version and their type and location inside the articles.
\begin{figure*}
    \centering
    \includegraphics[width=\textwidth]{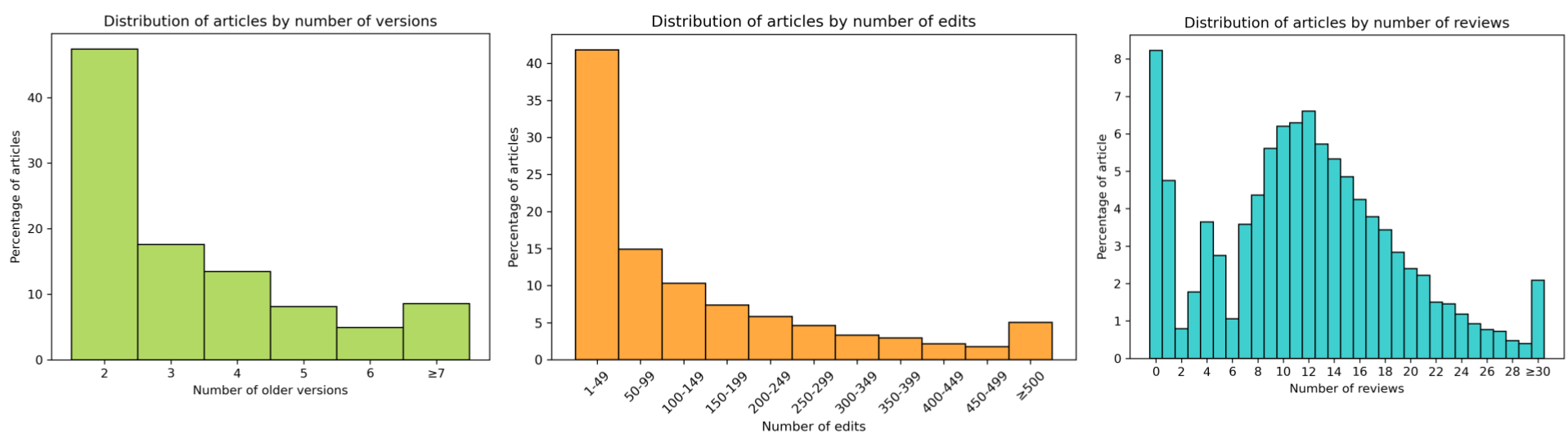}
    \caption{Distribution of articles by number of versions (left), edits (center) and reviews (right)}
    \label{fig:distri}
\end{figure*}

 \subsection{Distribution of versions and reviews}
In total, 390 GB of data was collected, comprising 121,492 PDFs for 29,504 articles, and their associated metadata (e.g.,~authors, venue, date of submission, keywords, etc.) and reviews.
After our creation process, our final corpus contains 36,733 pairs of versions distributed in 15,646 articles (one file is made of two successive versions of the same article, aligned sentence by sentence, where each pair of sentences has an associated list of edits if revised).
It encompasses contributions from 29 conferences (excluding independent submissions and challenges). The most represented domains are machine learning (ICLR, ICML, NeurIPS), robotics (RSS, CoRL), natural language processing (ACL), and computer vision (ECCV).

The distribution of the number of previous versions, edits, and reviews per article is depicted in Figure~\ref{fig:distri}. All articles have at least two versions, on average, each article has approximately 3.5 versions, allowing to consider the iterative aspect of the revision step. In terms of reviews, we considered all interactions within the article's forum, which explains the high variance in the number of reviews for specific articles. 

\subsection{Distribution of edits}

Table~\ref{table:edits} reports the distribution of both the length and the quantity of edits. Our corpus contains a total of 5.2M individuals edits in 3.7M edited sentences, with a wide variation of edit length and number of edits per articles.
To examine the intention behind these edits, we reported the distribution of the intention labels in our data in Table~\ref{table:intention}. This distribution is higher for \texttt{Content} and \texttt{Format} than in~\citet{jiang-etal-2022-arXivEdits}. This difference for the \texttt{Content} intention can be attributed to more substantial alterations, originating from having access to earlier versions: our data is collected from OpenReview rather than ArXiv where posted papers are closer to their final version. For the \texttt{Format} intentions, some errors remaining from the PDF conversion could be responsible for this difference since~\cite{jiang-etal-2022-arXivEdits} directly collected LaTeX files.

\begin{table}[htb!]
\begin{center}
\begin{tabularx}{\columnwidth}{|l|r||X|r|}

      \multicolumn{4}{c}{\textbf{Quantity of edits}}\\
      \hline
      Min & 1&First quartile&16\\
      \hline
      Max & 4432&Median&74\\
      \hline
     Average & 142.12&Third quartile&204\\\hline
     \multicolumn{4}{c}{\textbf{Edits length}}\\
     \hline
     Min&1     &Average&34.88\\\hline
     Max&9316 & Median &13\\
     \hline

\end{tabularx}
\caption{Distribution of the quantity of edits by articles and their length.}
\label{table:edits}
 \end{center} 
\end{table}

\begin{table}[htb!]
\begin{center}
\begin{tabularx}{\columnwidth}{|X|r|}

      \hline
      \textbf{Edit intention} & \textbf{Percentage}\\
      \hline
      Content & 41.97\%\\
      \hline
      Improve-grammar-typo & 22.73\%\\
      \hline
     Format & 20.38\%\\\hline
     Language &14.92\%\\
     \hline

\end{tabularx}

\caption{Distribution of edit intentions}
\label{table:intention}
 \end{center}
 
\end{table}

\subsection{Evolution and location of edits}
\begin{figure}[!ht]
\begin{center}
\includegraphics[scale=0.5]{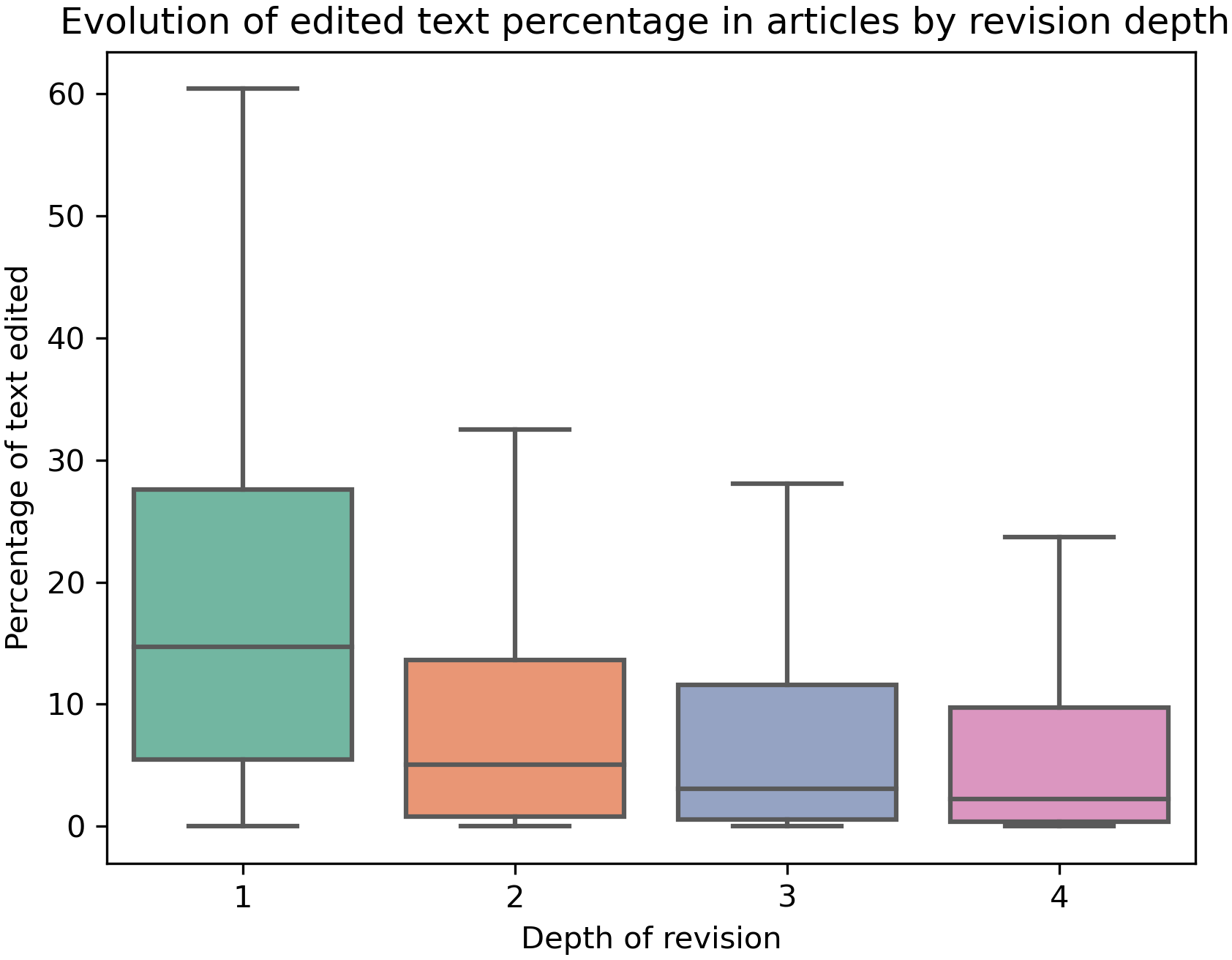} 
\caption{Evolution of edited text percentage in articles by revision depth.}
\label{fig:evolution_quantity}
\end{center}
\end{figure}
Figure~\ref{fig:evolution_quantity} shows the average percentage of the document revised for articles with 5 versions, resulting in 4 revisions, as a revision is the comparison of two successive versions. A trend emerges when we consider the percentage of text edited in an article by revisions' depth: an observable decrease in the extent of text revised as the depth of revisions increases.

\begin{figure}[!ht]
\begin{center}
\includegraphics[width=0.5\textwidth]{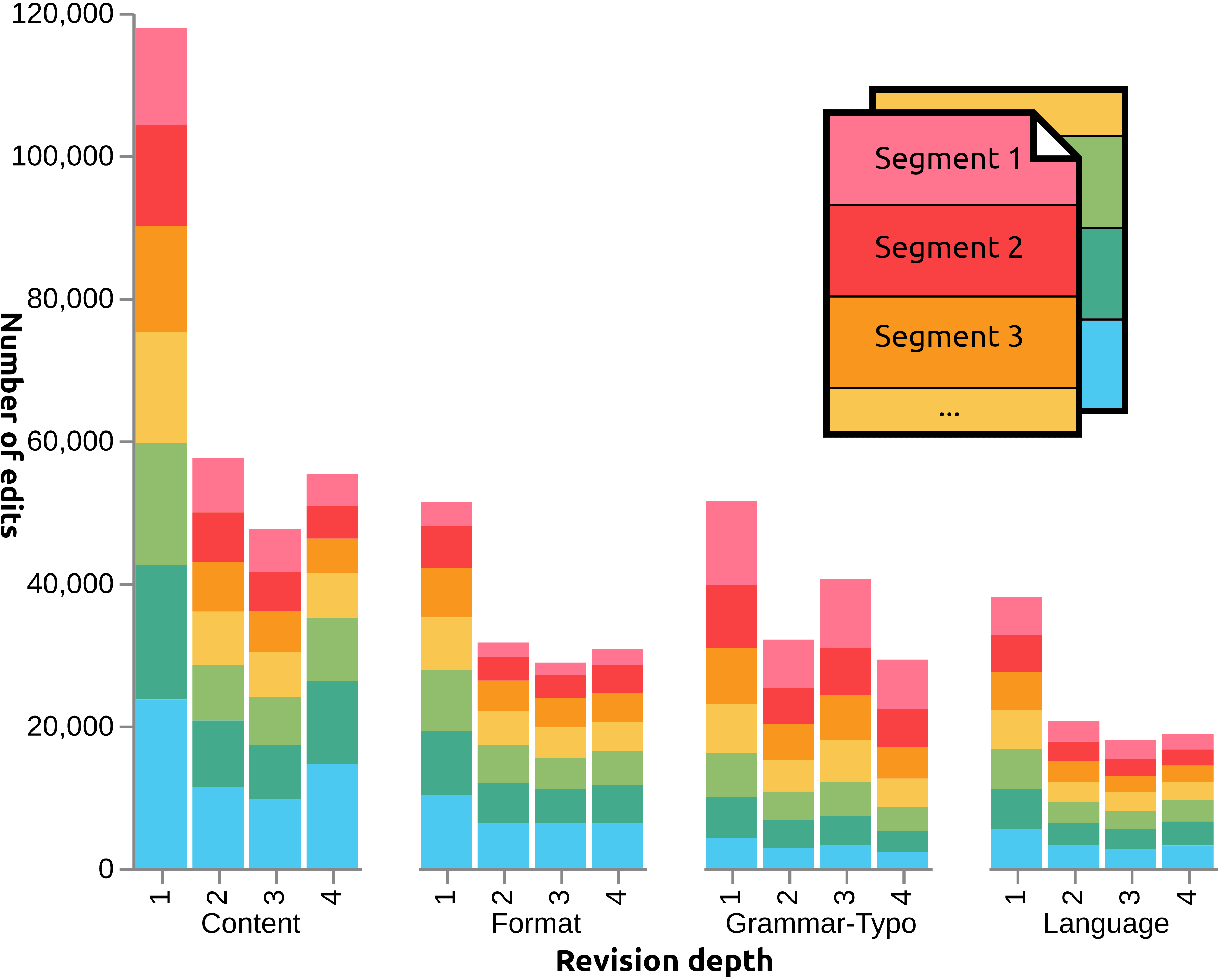} 
\caption{Evolution of the location of edited text by intention and revision depth}
\label{fig:evolution_location}
\end{center}
\end{figure}

Finally, we analysed the locations of revisions within the documents. Figure~\ref{fig:evolution_location} showcases the distribution of edits across document sections (evenly divided into 7 segments), categorized by edit intention and indexed by depth of revision, for articles with 5 versions. From this figure, we observe that \texttt{Content} tends to appear more towards the end of documents. This is not surprising as authors tend to add more content at the end of their document (adding new results, adding appendix, expanding limitations, writing acknowledgments, etc). We observe the same phenomenon with format edits. From our observations inside the documents, it seems to come from the addition of new figures and tables or changes in their placement. Grammar edits are more prevalent in the document's initial segments, it seems that authors are more thoughtful about their writing in the abstract and introduction sections. Language edits, in contrast, exhibit a uniform distribution throughout the document.

\section{Experiments with Text Revision Models}

One of the primary objectives of this corpus is to serve as a valuable resource for the task of text revision in scientific articles. Consequently, we evaluate some state-of-the-art models on our test data using various metrics frequently used to evaluate this task. We also explore the potential benefits of including Bertscore in this set of metrics for evaluating text revision from a semantic-based perspective.

\subsection{Baselines}
\label{s:baselines}

When selecting models for evaluation, we applied several criteria. First, we exclusively opted for open-source models. Consequently, despite previous research indicating the high efficiency of GPT models for this task~\citep{dwivediyu2022editeval}, we have chosen not to evaluate them. Instead, we selected a model that has made a significant impact on the field of text revision (Iterater), a state-of-the-art model specialized for this task (CoEdIT), and a state-of-the-art general-domain Large Language Model (LLM) (Llama2).

We compare the results of those baselines with the scores obtained when no revisions are applied, where the unrevised sentence is presented as the revised sentence without any alterations. We refer to this control approach as the \textbf{CopyInput} approach.

\paragraph{IteraTeR-PEGASUS}
\href{https://huggingface.co/wanyu/IteraTeR-PEGASUS-Revision-Generator}{\texttt{IteraTeR-PEGASUS}}\footnote{\href{https://huggingface.co/wanyu/IteraTeR-PEGASUS-Revision-Generator}{https://huggingface.co/wanyu/IteraTeR-PEGASUS-Revision-Generator}} is a fine-tuned version of \texttt{PEGASUS-LARGE} designed for the task of iterative text revision~\citep{du-etal-2022-read}.
To make inference possible,  we mapped categories from the ARXIVEDITS taxonomy to the Iterater taxonomy as follows:
"Improve-grammar-Typo" became "fluency", "Language" was categorized as both "clarity" and "coherence" \footnote{Due to their similar distribution in the Iterater corpus, we kept the two labels instead of choosing the most frequent one} and "Content" was labeled as "meaning-changed".

We explored two approaches for sentences with multiple revision intentions:
\begin{itemize}

    \item \texttt{best intention:} Treating the intentions separately and consistently providing the same intention at each iteration. This results in having $n$ revised sentences for an input sentence with $n$ intentions, from which we select the one with the maximum score.
    \item \texttt{all intentions:} Treating all intentions simultaneously, with a different intention given at each iteration. For a sentence with $n$ revision intentions, we force it to undergo at least $n$ iterations, resulting in a single output sentence for each input.
\end{itemize}

\paragraph{CoEdIT(XL)}
CoEdIT models are fine-tuned Flan-T5 models using the CoEdIT dataset~\citep{raheja2023coedit}.
We opted for \href{https://huggingface.co/grammarly/coedit-xl}{CoEdIT(XL)}\footnote{\href{https://huggingface.co/grammarly/coedit-xl}{https://huggingface.co/grammarly/coedit-xl}} due to its close performance to CoEdIT(XXL) but with nearly four times fewer parameters (11B for XXL compared to 3B for XL).
We selected CoEdIT as our specialized state-of-the-art model instead of PEER (chosen by~\citep{dwivediyu2022editeval}) because both are T5-based, but CoEdIT is the most recent and higher-performing option, as indicated by their results~\citep{raheja2023coedit}.

Similar to the approach used for Iterater, we experimented with two methods for sentences with multiple intentions:
\begin{itemize}
    \item \texttt{best intention:} We generated $n$ different revised sentences for a sentence with $n$ revision intentions and selected the one with the highest score according to the current metric.
    \item \texttt{all intentions:} We iteratively revised the sentence $n$ times with the $n$ different intentions.
\end{itemize}
 CoEdIT uses the same intention categories as~\cite{du-etal-2022-read} and requires specific prompts for each intention label. For each of our intention labels, we used the following prompts:
 \begin{itemize}
     \item \texttt{Improve-grammar-Typo}: 'Fix grammar errors in this sentence',
    \item \texttt{Language}: 'Clarify the sentence' and 'Improve the cohesiveness of the text',
    \item \texttt{Content}: 'Rewrite this sentence'
 \end{itemize}
 Notably, prompting for \texttt{Content} edits proved to be more challenging. From manual observations, a significant number of these edits involve substantial sentence rewriting rather than introducing new information to the text.
 
\paragraph{Llama2-7B}

Llama2 models are LLM released by Meta in July 2023. 
Due to hardware limitations, we could only run \href{https://huggingface.co/meta-llama/Llama-2-7b}{Llama2-7B}\footnote{\href{https://huggingface.co/meta-llama/Llama-2-7b}{https://huggingface.co/meta-llama/Llama-2-7b}} on our smaller test dataset. We employed the \texttt{best intention} and \texttt{all intentions} approaches similarly to what we did with the CoEdIT model. The only difference is that we added
\texttt{"\textbackslash n Corrected sentence: "} at the end of every prompt. Example: \texttt{"Clarify the sentence: <initial sentence> \textbackslash n Corrected sentence: "}

\paragraph{}
All baselines are evaluated on the task of sentence revision: \textquote[\citealp{jourdan2023text}]{the transformation of an input text into an improved
version fitting a desired attribute (formality, clarity, etc.), closer to the intended text.}. The inference is conducted on the 10\% and 3\% test split of our dataset. Our large test encompasses 3,733 pairs of documents for 1,597 articles and our small test encompasses 1,062 pairs of documents for 468 articles.
In our evaluation, we do not consider edits with \texttt{Format} intention, nor the insertion or deletion of entire sentences. This result in a set of 178K sentences to revise for the large test and 51K sentences for the small test.

\subsection{Metrics}
To evaluate the selected models, we employ five metrics. The choice of these metrics has been significantly influenced by the work of~\citet{du-etal-2022-read} and~\citep{dwivediyu2022editeval}. "References" refer to the actual revised sentences by the authors, extracted from the second version of a pair of articles. "Generated  sentences" refer to the revised sentences obtained by running the models on initial sentences.

    \paragraph{Exact-match (EM)} measures the rate of generated sentences that exactly match the references. While it is not the optimal metric due to its strict criteria, as even slight differences from the references result in a zero score, we use it for consistency with prior research.
    
    \paragraph{SARI}~\citep{xu-etal-2016-optimizing} is commonly employed in the evaluation of text revision, although it was originally designed for assessing automatic text simplification systems. SARI compares the system's output against both the references and the input sentence. It rewards the correct addition, deletion, and retention of words by the model. SARI is computed using the formula: $SARI=\frac{F1_{add} + F1_{keep} + P_{del}}{3}$ where F1 and P represent n-grams F1 score and n-grams precision with $n=4$. 
 
    \paragraph{BLEU}(Bilingual Evaluation Understudy) ~\cite{papineni-etal-2002-bleu} was originally developed for machine translation but has found use in various other tasks, including text revision. It is an n-gram-based metric that quantifies the similarity between the generated text and the reference text. A higher BLEU score indicates greater similarity between the two texts. %
    
    \paragraph{ROUGE-L}(Recall-Oriented Understudy for Gisting Evaluation)~\cite{lin-2004-rouge} is part of the ROUGE metrics, initially designed for evaluating automatic summarization. Like BLEU, ROUGE-L is an n-gram-based metric that measures the similarity between the generated and the reference texts written by humans. It measures the overlap in n-grams in terms of the longest common subsequence (LCS) between the reference and the generated texts. 
    \paragraph{Bertscore}~\cite{zhang2020bertscore} computes a cosine similarity score for matched words in reference and generated sentences using contextual embeddings from BERT. To the best of our knowledge, Bertscore has not been previously used for evaluating text revision. We include it in our evaluation because, unlike the other metrics, it should better capture the semantic meaning of sentences.

\subsection{Results}
\begin{table*}[!ht]
\begin{center}
\begin{tabularx}{\textwidth}{l X X X X X }

      \hline
      \textbf{Model/Metric}&EM&BLEU&ROUGE&SARI&BERT\\
      \hline
      CopyInput & 0.00&\textbf{66.31}&\textbf{74.19}&61.38 &94.46\\
      \hline
      Iterater-Pegasus (best intention) & 6.04 & 60.99 & 73.25 & 55.27&95.93\\
      Iterater-Pegasus (all intentions) & 5.98 & 58.68 & 72.36 & 53.77 & 93.29 \\
      \hline
     CoEdIT (best intention) & 8.27&58.88&70.89&53.94&\textbf{96.08}\\
     CoEdIT (all intentions) & 8.25& 56.44 & 69.22& 51.62 & 95.99\\
          \hline
     Llama2-7B (best intention)\ding{168} & \textbf{14.05} & 61.91 & 73.02 & \textbf{62.07} & 92.84\\
    Llama2-7B (all intentions) \ding{168} & 13.76 & 57.46 & 68.18 & 58.39 & 92.37 \\
      \hline

\end{tabularx}
\caption{Results for all baselines. \ding{168} are results on the small set, others are realized on the large set.}
\label{table:resulteval}
 \end{center}
\end{table*}
We report the results of our experiments in Table~\ref{table:resulteval}. Due to material limitations, Llama2-7B was only evaluated on the small test. Iterater-Pegasus and CoEdIT were evaluated on both the large and small tests. For brevity, we only provide the results from the large test here, as the results from the small test were consistent.

Among the various approaches, Llama-7B (best intention) and CopyInput give the best performances.
When considering the two tools based on LM, Iterater-Pegasus, despite being older, outperforms CoEdIT on conventional metrics.
However, when evaluated with Bertscore, CoEdIT consistently holds a slight edge. The use of Bertscore revealed a different model ranking than other metrics, although all approaches achieved high scores using this metric.
Overall, across all metrics, the approaches yield closely matched results. This observation, coupled with the good performance of CopyInput in comparison to other methods, lead us to question the current evaluation methods.

The issues with the evaluation methodology seem to stem from the 1-to-n nature of the text revision task. Traditional evaluation methods involve comparing the predicted revised sentence to the actual sentence modification. However, there may be alternative and potentially superior revisions of a sentence far from the gold revision that will, in consequence, obtain a low score with the currently used metrics.
One of the challenges in evaluating text revision is to establish an evaluation approach that genuinely reflects the models' quality in performing this task. A promising direction for text revision evaluation involves experimenting with an aggregation of metrics that go beyond the comparison of the initial sentence to the target sentence (improvement in grammaticality~\citep{choshen-abend-2018-automatic} or readability~\citep{chall1995readability,cohmetrix}). Another direction could be to use multiple ground truth revision, either produced manually (preferably) or generated automatically using paraphrase systems.

\section{Conclusion}

In this article, we introduced \texttt{CASIMIR}, the largest corpus for scientific text revision, and provided a detailed description of its creation process. We conducted a qualitative analysis of its content and evaluated the performance of baseline text revision tools on our test split. We used a set of commonly employed metrics for this task and introduced an existing semantic metric, Bertscore, which was originally applied in this work to text revision.

Experiments revealed that state-of-the-art approaches failed to surpass our control approach (CopyInput) on the majority of metrics. While Llama-7B emerged as the top-performing model, the results were so closely matched that drawing significant conclusions proved challenging. These findings have prompted us to question the effectiveness of the current evaluation approach for the task of text revision.

Throughout this work, we encountered several challenges. One of the main challenges was related to PDF conversion. Currently, it remains an unresolved challenge, although ongoing research in the field is leading to the release of new tools~\cite{shen-etal-2022-vila,blecher2023nougat}. Another challenge arose from the alignment and edit extraction process, as there is no all-in-one tool available for these tasks and most libraries do not offer word-level diff extraction.

Our dataset is freely available\footnote{\href{https://huggingface.co/datasets/taln-ls2n/CASIMIR}{https://huggingface.co/datasets/taln-ls2n/CASIMIR}}. It can be employed for training text revision models capable of considering contextual information beyond individual sentences. Moreover, the incorporation of peer reviews opens up possibilities for diverse applications, including predicting paper acceptance/rejection, automating review generation, and automated text quality assessment.

 \section{Limitations and Ethical Considerations }
We used a variety of automatic tools during our process. \citealp{shen-etal-2022-vila} reported a performance of 83.77\% on their dataset for PDF extraction, \citealp{bertalign} reported a performance of 99\% for alignment and \citealp{jiang-etal-2022-arXivEdits} 84.4\% for intent classification. However, we do not have any information on the performances of \texttt{git-diff} used for edit extraction.
Errors made by these tools persist throughout the entire dataset creation process, introducing additional noise, as no large-scale manual checking has been done. It is important to consider this when using the data for future training.

All the data in our dataset was collected from publicly available sources. Articles on OpenReview fall under different "non-exclusive, perpetual, and royalty-free license"\footnote{\href{https://openreview.net/legal/terms}{https://openreview.net/legal/terms}}, and reviews are licensed under CC BY 4.0.
Our dataset exclusively comprises scientific articles and their associated comments. However, since we did not manually review each document, we cannot guarantee that it contains no personal data provided by authors or hate speech. This is especially relevant in the case of review comments, as OpenReview is entirely open, allowing users to freely express their opinions. Individuals interested in training a model on this dataset should take these considerations into account.

 \section{Acknowledgments}
This work was granted access to the HPC resources of IDRIS under the allocation 2022-AD011013901 made by GENCI. \\
This research was funded, in whole or in part, by l’Agence Nationale de la Recherche (ANR), project ANR-22-CE38-0004.
\section{Bibliographical References}\label{sec:reference}

\bibliographystyle{lrec-coling2024-natbib}
\bibliography{lrec-coling2024-example}


\end{document}